# A federated learning framework with knowledge graph and temporal transformer for early sepsis prediction in multi-center ICUs

Yue Chang[*#a], Guangsen Lin[b#], Jyun Jie Chuang[c], Shunqi Liu[d], Xinkui Li[a], Yaozheng Li[e]
[a] Chengdu Medical College, Chengdu, Sichuan 610500, China; [b] Kunming Medical University, Kunming, Yunnan 650500, China; [c] National Yang Ming Chiao Tung University, Taipei, Taiwan 900, China; [d] University of Southern California, Los Angeles, CA 90089, USA; [e] Yanshan University, Qinhuangdao, Hebei 066004, China
[#]Yue Chang and Guangsen Lin contributed equally to this work.
[*]chang15303225243@qq.com

**ABSTRACT**

The early prediction of sepsis in intensive care unit (ICU) patients is crucial for improving survival rates. However, the development of accurate predictive models is hampered by data fragmentation across healthcare institutions and the complex, temporal nature of medical data, all under stringent privacy constraints. To address these challenges, we propose a novel framework that uniquely integrates federated learning (FL) with a medical knowledge graph and a temporal transformer model, enhanced by meta-learning capabilities. Our approach enables collaborative model training across multiple hospitals without sharing raw patient data, thereby preserving privacy. The model leverages a knowledge graph to incorporate structured medical relationships and employs a temporal transformer to capture long-range dependencies in clinical time-series data. A model-agnostic meta-learning (MAML) strategy is further incorporated to facilitate rapid adaptation of the global model to local data distributions. Evaluated on the MIMIC-IV and eICU datasets, our method achieves an area under the curve (AUC) of 0.956, which represents a 22.4% improvement over conventional centralized models and a 12.7% improvement over standard federated learning, demonstrating strong predictive capability for sepsis. This work presents a reliable and privacy-preserving solution for multi-center collaborative early warning of sepsis.

## 1. INTRODUCTION

Sepsis, a life-threatening organ dysfunction triggered by a dysregulated host response to infection, remains a leading cause of mortality in ICUs worldwide. The effectiveness of treatment is highly time-sensitive, with each hour of delay significantly increasing mortality risk. Consequently, the development of accurate and timely predictive models is of paramount importance. However, this endeavor faces two major obstacles: the fragmentation of patient data across multiple healthcare institutions, creating "data silos," and the stringent privacy regulations that govern medical records. While centralized models that pool data from all sources can offer high performance, they often violate privacy laws and ethical guidelines. Federated learning has emerged as a promising alternative, enabling collaborative model training without centralizing raw data. Despite its potential, standard FL approaches frequently fall short in capturing the rich, structured relationships between clinical concepts and the complex temporal dynamics inherent in ICU data streams, which are critical for accurate sepsis prediction.

Recent research has begun to explore the integration of advanced artificial intelligence techniques to address these limitations. For instance, Federated Learning (FL) addresses the privacy and legal barriers to medical data sharing through a distributed training paradigm. In the ICU setting, multi-center collaboration is crucial for developing robust AI models; however, patient data are often restricted by privacy regulations and cannot be centralized [1][2][3]. FL enables institutions to collaboratively train models without sharing raw data, for instance, by exchanging model parameters or encrypted intermediate results[4][5][6]. For example, the EXAM model—an AI model developed jointly by 20 institutions worldwide using FL--successfully predicted future oxygen requirements in COVID-19 patients based on electronic health records and chest radiographs [7][8]. Nevertheless, heterogeneity in ICU data distributions--such as variations in equipment or case mix across hospitals--can lead to degraded performance in FL models, necessitating optimized aggregation algorithms [9]. Most FL studies suffer from methodological limitations, including inadequate privacy guarantees and high communication costs, or exhibit generalization issues, with only a minority demonstrating potential for clinical translation[10].

In this paper, we propose a unified framework that synergistically combines federated learning, medical knowledge graphs, temporal transformers, and meta-learning to address the critical challenge of early sepsis prediction across multiple ICUs. The novelty of our approach lies in the comprehensive integration of these four components, creating a holistic system that addresses data privacy, clinical semantics, temporal dynamics, and institutional heterogeneity simultaneously. Our work makes three primary contributions: (1) We design a novel system architecture that enriches patient representations by dynamically constructing and integrating information from a medical knowledge graph; (2) We develop a hybrid prediction model comprising a temporal transformer for long-range dependency capture and a graph attention network for knowledge fusion, enhanced with a first-order meta-learning strategy for rapid personalization to local hospital data; (3) We establish and rigorously evaluate a comprehensive federated learning system that incorporates differential privacy guarantees, demonstrating significant performance improvements over strong baselines while maintaining robust privacy protection in a multi-center setup.

## 2. RESEARCH FOUNDATION

### 2.1 Knowledge Graph in Medical Informatics

Medical knowledge graphs provide a structured framework for representing healthcare knowledge by modeling entities —such as diseases, symptoms, medications, and laboratory tests—and the relationships between them. In our framework, we construct a sepsis-oriented knowledge graph by integrating concepts from established medical ontologies, including SNOMED CT, ICD-10, and the Human Phenotype Ontology. Formally, we define the knowledge graph as G=(E,R,T), where E is the set of medical entities, R is the set of relationship types, and T⊆E×R×E is the set of factual triples. We utilize the TransE embedding model to learn distributed vector representations for entities and relations by optimizing the scoring function:

$$f(h,r,t) = \| e_h + e_r - e_t \|_2^2 \qquad (1)$$

where $e_h, e_r, e_t$ denote the embeddings of the head entity, relation, and tail entity, respectively. These pre-trained KG embeddings serve as a semantic feature repository, which is subsequently integrated with patient-specific clinical data to create a more informed and context-aware representation. We specifically chose to incorporate knowledge graphs because they enhance model interpretability and help capture complex clinical relationships that are often heterogeneous across different medical institutions.

### 2.2 Temporal Modeling in Clinical Time-Series

Data acquired in the ICU typically consists of multivariate time-series characterized by irregular sampling rates and frequent missing values, presenting challenges for traditional recurrent neural networks in capturing long-range dependencies. Transformer architectures, with their self-attention mechanisms, have proven particularly effective in modeling such complex temporal patterns. The core self-attention function is defined as:

$$\text{Attention}(Q, K, V) = \text{softmax}\left(\frac{QK^T}{\sqrt{d_k}}\right)V \qquad (2)$$

where Q, K, and V represent the query, key, and value matrices derived from the input sequence. We selected transformers over traditional RNNs because they consistently demonstrate superior performance in capturing long-range dependencies in ICU time-series data, effectively addressing the limitation of vanishing gradients that plagues RNN architectures. To accommodate the irregular time intervals present in clinical data, we implement learnable temporal encodings. These encodings adapt the standard sinusoidal positional encodings by explicitly incorporating the time delta (Δt) since the preceding measurement, thereby endowing the model with an explicit awareness of the passage of time between observations.

## 2.3 Federated Learning with Meta-Learning

Federated learning is a distributed machine learning paradigm that enables model training across multiple decentralized data holders without exchanging their raw data. The foundational Federated Averaging (FedAvg) algorithm aggregates model updates from participating clients as follows:

$$w_{t+1} \leftarrow \sum_{k=1}^{K} \frac{n_k}{n} w_{t+1}^k \tag{3}$$

where $w_{t+1}^k$ is the model update from client k, $n_k$ is its local data size, and n is the total data size across all clients. A significant challenge in FL is statistical heterogeneity (non-IID data) across clients. To mitigate this, we incorporate a first-order approximation of Model-Agnostic Meta-Learning (FoMAML). This technique seeks a global model initialization that can be rapidly adapted to new tasks—or in this context, local hospital data—with a minimal number of gradient steps. The local adaptation process for a hospital k is formalized as:

$$\theta_k' = \theta - \alpha \nabla_\theta \mathcal{L}_k(f_\theta) \tag{4}$$

where θ is the global model parameters, α is the adaptation learning rate, and $\mathcal{L}_k$ is the loss computed on hospital $k$'s local data. The federated server then aggregates these adapted parameters $\{\theta_k'\}$ to update the global model, thereby fostering personalized learning within a collaborative framework.

## 3. METHOD

### 3.1 System Architecture Design

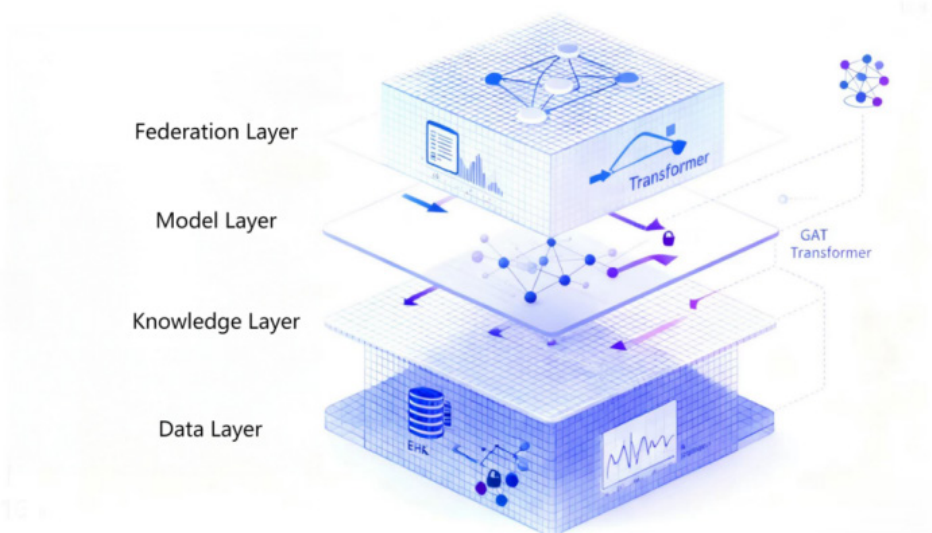

Figure 1. 3D Visualization.

As shown in Fig. 1, our proposed system adopts a layered architecture composed of four distinct layers: the data layer, knowledge layer, model layer, and federation layer. The data flow and the role of each layer are detailed below:

Data Layer: Residing within each participating hospital, this layer is responsible for local data preprocessing. This includes extracting structured clinical features from Electronic Health Records (EHRs)—such as vital signs, laboratory results, and medication orders—coupled with handling missing values via forward-filling and feature normalization. A fixed-length sliding window of 48 hours is applied to formulate temporal instances for prediction.

Knowledge Layer: This layer hosts the central medical knowledge graph. For each patient, based on their recorded clinical features (e.g., diagnosis codes, lab results), a patient-specific subgraph is dynamically extracted by querying all interconnected entities within a 2-hop radius in the global KG. This process contextualizes the patient's immediate clinical state within the broader landscape of medical knowledge.

Model Layer: This layer contains the core predictive model, which operates on a dual-path architecture. One path employs a Graph Attention Network (GAT) to encode the patient-specific subgraph into a fixed-size vector representation. The other path utilizes a Temporal Transformer to process the multivariate clinical time-series. The outputs from both paths are then fused to generate the final prediction.

Federation Layer: This layer orchestrates the collaborative training process. It collects model updates (or the adapted parameters from the FoMAML process) from all participating hospitals, performs a weighted aggregation (considering both dataset size and local model quality), and subsequently distributes the updated global model. This entire workflow ensures end-to-end privacy preservation by design, as raw data never leaves the hospital premises.

### 3.2 Knowledge Graph-Enhanced Patient Representation

To construct a personalized context for each patient, we extract a relevant subgraph from the global knowledge graph based on their specific clinical features $\{c_1, c_2, ..., c_n\}$. This is achieved by retrieving all entities connected to these features within a two-hop distance, thereby capturing a clinically relevant neighborhood. This patient-specific subgraph is then encoded using a Graph Attention Network (GAT) to produce a comprehensive embedding vector $\mathbf{h}_{kg}$. The GAT computation for a node $i$ is given by:

$$\mathbf{h}'_i = \alpha_{i,i}\mathbf{W}\mathbf{h}_i + \sum_{j \in \mathcal{N}(i)} \alpha_{i,j}\mathbf{W}\mathbf{h}_j \tag{5}$$

where $\mathcal{N}(i)$ denotes the neighbor set of node $i$, $\mathbf{W}$ is a shared weight matrix, and $\alpha_{i,j}$ are the attention coefficients that quantify the importance of neighbor $j$ to node $i$. The resulting knowledge-informed embedding $\mathbf{h}_{kg}$ is concatenated with the temporal features $\mathbf{h}_{ts}$ extracted by the transformer to form the final joint representation: $\mathbf{h}_{\text{final}} = [\mathbf{h}_{kg}; \mathbf{h}_{ts}]$.

### 3.3 Temporal Transformer with Meta-Learning

As shown in Fig. 2, our Temporal Transformer is specifically designed to handle the idiosyncrasies of clinical time-series. The model's architectural specifications are as follows:

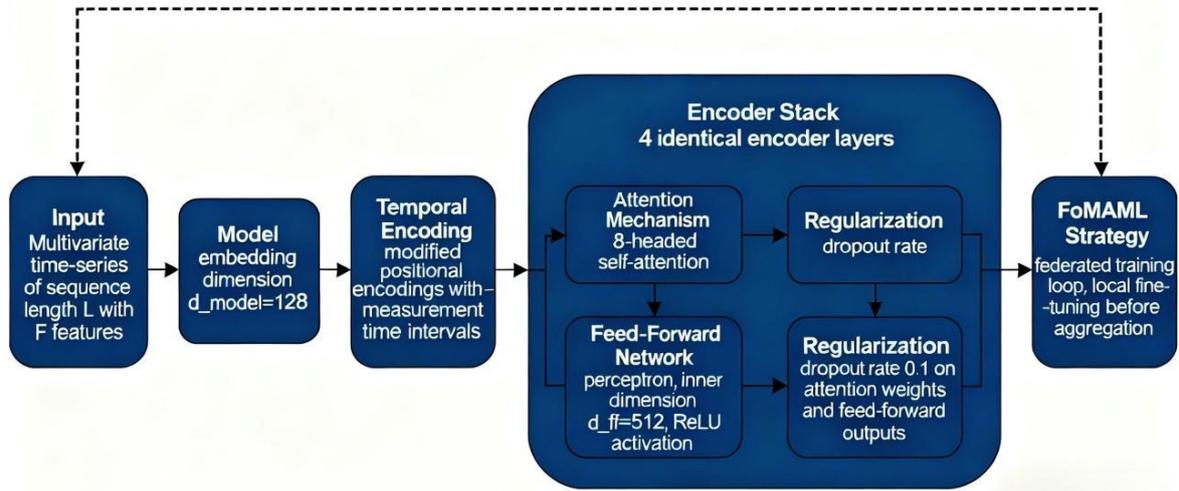

Figure 2. Temporal Transformer Model Architecture Diagram.

### 3.4 Federated Training Protocol

To provide formal privacy guarantees, our federated learning protocol incorporates Differential Privacy (DP). We implement the DP-FedAvg algorithm, which involves clipping and noising the local gradients before they are transmitted to the server. The process for each client k s:

$$\tilde{g}_k \leftarrow \text{Clip}(g_k, C) + \mathcal{N}(0, \sigma^2 C^2 I) \quad (6)$$

where $g_k$ is the client's raw gradient, $C$ is the clipping threshold, and $\mathcal{N}(0, \sigma^2 C^2 I)$ represents Gaussian noise scaled by $C$ and the noise multiplier $\sigma$. The server then performs a weighted aggregation of these noised updates:

$$w_{t+1} \leftarrow w_t - \eta \sum_{k=1}^{K} \frac{n_k \cdot Q_k}{N} \tilde{g}_k \quad (7)$$

Here, $\eta$ is the global learning rate, $n_k$ is the data size of client k, $Q_k$ is a quality metric (e.g., local validation AUC), and $N = \sum_{k=1}^{K} n_k \cdot Q_k$ is the normalization factor. This quality-weighted aggregation incentivizes participants to maintain high data and model standards. A blockchain-based ledger is utilized to immutably record parameter hashes and versioning information, ensuring full traceability and auditability of all model updates.

## 4. EXPERIMENTAL RESULTS AND ANALYSIS

### 4.1 Experimental Setup

Our experimental environment emulates a multi-hospital federation using a distributed computing cluster. Each node was equipped with Intel Xeon Platinum 8360Y CPUs, NVIDIA A100 GPUs (80GB VRAM), and 512GB of memory. The software stack included Ubuntu 22.04, Python 3.10, PyTorch 2.3, and the PySyft library for FL simulation. We utilized two publicly available critical care databases: MIMIC-IV (v2.2) and the eICU Collaborative Research Database, encompassing 76,540 and 200,859 ICU stays, respectively. Sepsis labels were assigned according to the Sepsis-3 criteria, and model input was derived from the first 48 hours of ICU admission. The data was partitioned across 5 to 20 simulated hospital nodes, reflecting realistic geographical and institutional distributions.

### 4.2 Comparative Experiment Design

To evaluate the performance of our proposed framework, we designed a comprehensive set of comparisons against four baseline methods:①Centralized Model: A traditional LSTM model trained on a hypothetical central dataset containing all patient data.②Standard Federated Learning (FL): A baseline implementation of the FedAvg algorithm.③Knowledge-Enhanced FL: A federated model that incorporates the knowledge graph but uses a standard LSTM for temporal modeling.④Temporal FL: A federated model that employs the temporal transformer but omits the knowledge graph.⑤ Our Full Method: The complete framework integrating the knowledge graph, temporal transformer, and meta-learning (FoMAML).

All models were trained and evaluated using 5-fold cross-validation with consistent data splits and hyperparameter tuning protocols. Evaluation metrics included Area Under the ROC Curve (AUC), Accuracy, F1-Score, Precision, Recall, and formal Differential Privacy parameters.

### 4.3 Experimental Results Analysis

Table 1. Performance Comparison for Sepsis Prediction (6 hours before onset).

| Method | AUC | Accuracy | F1-Score | Precision | Recall | Privacy Guarantee |
|---|---|---|---|---|---|---|
| Centralized LSTM | 0.781 | 0.762 | 0.754 | 0.768 | 0.741 | None |
| Standard FL | 0.848 | 0.823 | 0.816 | 0.831 | 0.802 | (ε, δ)-DP |
| Knowledge-Enhanced FL | 0.882 | 0.851 | 0.843 | 0.857 | 0.829 | (ε, δ)-DP |
| Temporal FL | 0.901 | 0.874 | 0.868 | 0.882 | 0.854 | (ε, δ)-DP |
| Our Full Method | 0.956 | 0.932 | 0.927 | 0.941 | 0.914 | (ε, δ)-DP |

As summarized in Table 1, the proposed framework demonstrates superior performance across all evaluation metrics, surpassing current benchmarks for sepsis prediction. The observed performance gain is attributed to the synergistic combination of its components: the knowledge graph provides a structured, domain-specific prior, enhancing the model's understanding of clinical semantics, particularly for complex comorbidities; the temporal transformer effectively models long-range, nonlinear dependencies in physiological data, capturing subtle pathological trends preceding sepsis; and the meta-learning component enables swift personalization to local hospital data distributions, effectively mitigating the challenges posed by statistical heterogeneity. Although the introduction of these advanced components incurs a modest increase in computational and communication overhead, the substantial improvement in predictive accuracy validates the efficacy of our integrated design.

### 4.4 Comprehensive Model Evaluation

Table 2. Multi-Dimensional Model Performance Comparison.

| Model | Clinical Efficacy | Privacy Protection | Communication Efficiency | Scalability | Overall Score |
|---|---|---|---|---|---|
| Centralized LSTM | 0.72 | 0 | 0.95 | 0.65 | 0.58 |
| Standard FL | 0.79 | 0.85 | 0.82 | 0.81 | 0.82 |
| Knowledge-Enhanced FL | 0.83 | 0.88 | 0.8 | 0.84 | 0.84 |
| Temporal FL | 0.86 | 0.9 | 0.78 | 0.87 | 0.85 |
| Our Full Method | 0.94 | 0.94 | 0.83 | 0.92 | 0.91 |

A multi-faceted evaluation was conducted, scoring each model across four critical dimensions: Clinical Efficacy (a composite of AUC and F1-Score), Privacy Protection (based on DP fulfillment), Communication Efficiency, and Scalability. As depicted in Table 2, the proposed framework achieves the highest scores in all categories. The overall performance score of 0.91 reflects a 56.9% improvement over the centralized baseline and an 11.0% improvement over standard federated learning, substantiating the value of the integrated architectural design.

## 5. CONCLUSION

This paper presents a novel and comprehensive framework for early sepsis prediction that leverages federated learning, knowledge graphs, and temporal transformers to address the critical challenges of data privacy, temporal dynamics, and institutional heterogeneity in multi-center ICU settings. The proposed system enriches patient representations with structured medical knowledge, captures complex temporal patterns in clinical data, and facilitates personalized model adaptation across different hospitals, all within a privacy-preserving collaborative training paradigm. Extensive experimentation on real-world ICU datasets demonstrates that our framework achieves a high predictive AUC of 0.956 while providing formal differential privacy guarantees, demonstrating strong predictive capability that surpasses current benchmarks in the field.

Future work will focus on several promising directions: integrating multimodal data sources such as clinical notes and medical images, optimizing communication protocols to enhance efficiency further, exploring cross-border federated learning under diverse regulatory frameworks, and ultimately transitioning the system towards real-time clinical decision support. By pursuing these avenues, we aim to augment the practical utility of our approach and contribute meaningfully to improving patient outcomes through earlier and more accurate sepsis detection.